\documentclass[10pt,twocolumn,letterpaper]{article}

\usepackage{iccv}
\usepackage{times}
\usepackage{epsfig}
\usepackage{graphicx}
\usepackage{amsmath}
\usepackage{algorithmic}
\usepackage{algorithm}
\usepackage{amssymb}
\usepackage{bbm}
\usepackage{svg}
\usepackage{multirow}
\usepackage{amsmath}
\usepackage{makecell}
\usepackage{diagbox}
\usepackage{svg}
\usepackage{array}

\usepackage[pagebackref=true,breaklinks=true,letterpaper=true,colorlinks,bookmarks=false]{hyperref}

\iccvfinalcopy 

\def\iccvPaperID{7138} 

\ificcvfinal\fi

\begin{document}

\title{Distribution Shift Matters for Knowledge Distillation \\with Webly Collected Images}

\author{Jialiang Tang$^{1,2,3}$, Shuo Chen$^{4,*}$, Gang Niu$^{4}$, Masashi Sugiyama$^{4,5}$, Chen Gong$^{1,2,3,}$\thanks{Corresponding authors: Chen Gong (chen.gong@njust.edu.cn), Shuo Chen (shuo.chen.ya@riken.jp).} \\
$^{1}$School of Computer Science and Engineering, Nanjing University of Science and Technology, China \\ 
$^{2}$Key Laboratory of Intelligent Perception and Systems for\\ High-Dimensional Information of Ministry of Education, China \\
$^{3}$Jiangsu Key Laboratory of Image and Video Understanding for Social Security, China \\
$^{4}$Center for Advanced Intelligence Project, RIKEN, Japan \\
$^{5}$The Graduate School of Frontier Sciences, The University of Tokyo, Japan \\
}
\maketitle
\ificcvfinal\thispagestyle{empty}\fi

\begin{abstract}
Knowledge distillation aims to learn a lightweight student network from a pre-trained teacher network. In practice, existing knowledge distillation methods are usually infeasible when the original training data is unavailable due to some privacy issues and data management considerations. Therefore, data-free knowledge distillation approaches proposed to collect training instances from the Internet. However, most of them have ignored the common distribution shift between the instances from original training data and webly collected data, affecting the reliability of the trained student network. To solve this problem, we propose a novel method dubbed ``Knowledge Distillation between Different Distributions" (KD$^{3}$), which consists of three components. Specifically, we first dynamically select useful training instances from the webly collected data according to the combined predictions of teacher network and student network. Subsequently, we align both the weighted features and classifier parameters of the two networks for knowledge memorization. Meanwhile, we also build a new contrastive learning block called MixDistribution to generate perturbed data with a new distribution for instance alignment, so that the student network can further learn a distribution-invariant representation. Intensive experiments on various benchmark datasets demonstrate that our proposed KD$^{3}$ can outperform the state-of-the-art data-free knowledge distillation approaches.
\end{abstract}

\section{Introduction}
\label{sec_intro}
In recent years, advanced deep neural networks (DNNs) have significantly succeeded in many computer vision fields~\cite{girshick2015fast,he2017mask}. However, those excellent DNNs usually have excess learning parameters, which may incur unaffordable computation and memory burdens for resource-limited intelligent devices. To address this problem, model compression algorithms have been developed to constrict heavy DNNs into portable ones, mainly including the network pruning~\cite{liu2017learning}, network quantization~\cite{rastegari2016xnor}, and knowledge distillation~\cite{hinton2015distilling}. 

Most existing compression algorithms are data-driven and rely on massive original training data that is usually inaccessible in the real world. For example, the large-scale ImageNet~\cite{deng2009imagenet} requires 138\textit{GB} of storage and is too heavy to transfer among devices, yet the ResNet34~\cite{he2016deep} trained on ImageNet only needs 85\textit{MB} memory and can be shared at a relatively low cost. Besides, users may be more willing to share pre-trained models than their personal data, such as photos and travel records. As a result, existing data-driven algorithms for model compression frequently fail to deal with large DNNs in practical applications.  
\begin{figure}[t]
	\centering
	\includegraphics[scale=0.13]{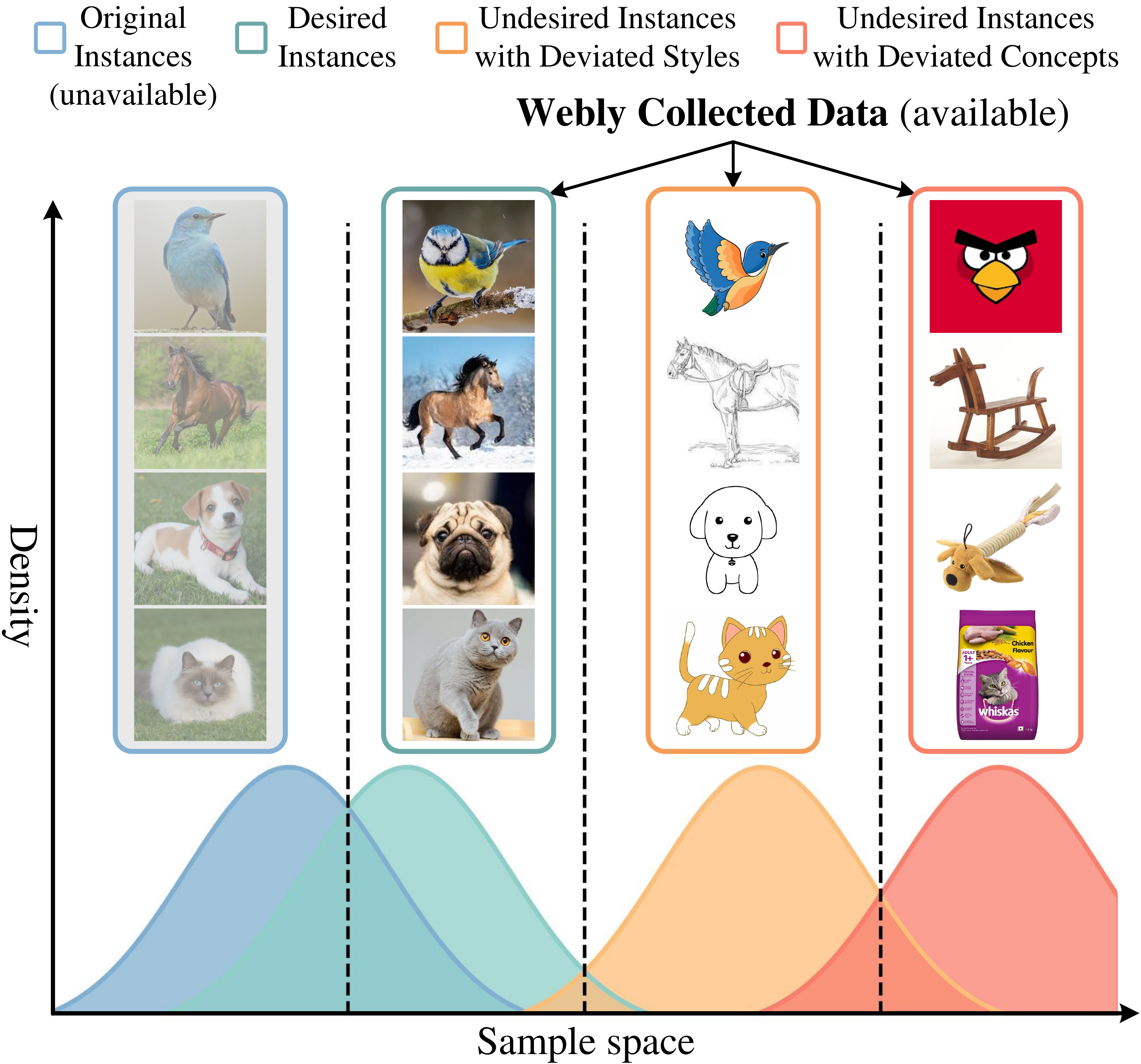}
	\caption{The illustration of distribution shift between the webly collected data and original data, where the original data consists of realistic images of animals. Nevertheless, the webly collected data may include cartoon and sketch images of animals, and even some non-animal images.}
	\label{fig_ds}
\vspace{-1.5em}
\end{figure}

To address this issue, data-free model compression methods have received wide attention in recent studies~\cite{chen2021learning,choi2020data,do2022momentum,fang2021contrastive}. Among these methods, data-free knowledge distillation has shown encouraging results, which only requires a pre-trained large network (\textit{a.k.a.} a teacher network) to learn a compact network (\textit{a.k.a.} a student network). Existing data-free knowledge distillation methods train student network with the guidance of teacher network through the generated pseudo data~\cite{chen2019data,yin2020dreaming,zhao2022dual} or real-world data collected from the Internet~\cite{chen2021learning}. Generally, the performance of the student networks trained on synthetic data might be suboptimal due to the flawed or distorted synthetic images. In comparison, the student networks using real-world data from the Internet usually achieve better performance, especially on the tasks involving complicated natural images. 

Current data-free knowledge distillation methods~\cite{chen2021learning} that train student network with data from the Internet (\textit{i.e.}, webly collected data) seek to select confident instances from the collected data, so that they can provide correctly labeled images for training student network. However, the webly collected data and original data may have different distributions, and existing methods usually ignore the distribution shift (\textit{e.g.}, the image style and image category) between them, as shown in Fig.~\ref{fig_ds}. For example, when we are interested in classifying various real-world animals and enter ``cat" into the image search engines, we may obtain the images of ``cartoon cat" or ``cat food". Apparently, the former is with different styles of cat images, and the latter is even unrelated to our interested animal classification task. The student network trained on the webly collected data will inevitably suffer from distribution shift when it is evaluated on the unseen test data. This makes the performance of student network trained on the webly collected data obviously lower than that using the original data. Consequently, it is critically important to alleviate the distribution shift between the webly collected data and original data.

To this end, we propose a new data-free approach called \textbf{K}nowledge \textbf{D}istillation between \textbf{D}ifferent \textbf{D}istributions (KD$^{3}$) to learn a student network by utilizing the plentiful data collected from the Internet with specific considerations on the distribution shift. More specifically, we first select the webly collected instances with the similar distribution to original data by dynamically combining the predictions of teacher network and student network during the training phase. After that, to exhaustively learn the information of teacher network, we share the classifier of teacher network with student network and conduct a weighted feature alignment. In this way, we can encourage student network to mimic the feature extraction of teacher network. Furthermore, a new contrastive learning block MixDistribution is designed to control the statistics (\textit{i.e.}, the mean and variance) of instances, so that we can generate perturbed instances with the new distribution. The student network is encouraged to produce consistent features for the unperturbed and perturbed instances to learn the distribution-invariant representation, which can generalize to the previously unseen test data. As a result, the student network that precisely mimics teacher network can produce the features that are consistent with teacher network. Finally, these features fed into the shared classifier can make the predictions as accurate as the corresponding teacher network. Thanks to effectively resolving the distribution shift between the webly collected data and original data, our KD$^{3}$ finally learns an accurate and lightweight student network, which can achieve comparable performance to those student networks trained on the original data. The contributions of our proposed KD$^{3}$ are summarized as follows:
\begin{itemize}
	\item We propose a new data-free knowledge distillation method termed KD$^{3}$, which dynamically selects useful training instances from the Internet by alleviating the distribution shift between the original data and webly collected data.
	\item We design a weighted feature alignment strategy and a new contrastive learning block to closely match student network with teacher network in the feature space, so that the student network can successfully learn useful knowledge from teacher network for the unseen original data.
	\item Intensive experiments on multiple benchmarks demonstrate that our KD$^{3}$ can outperform the state-of-the-art data-free knowledge distillation approaches. 
\end{itemize}
\section{Related Works}
\label{sec_rw}
In this section, we review previous works related to our proposed KD$^{3}$, mainly including knowledge distillation and the learning approaches under distribution shift.
\subsection{Knowledge Distillation}
Conventional knowledge distillation usually needs the original training data to launch knowledge transfer from a teacher to a student. In general, they utilize the soften predictions~\cite{ba2014deep,hinton2015distilling}, middle-layer features~\cite{chen2022knowledge,romero2014fitnets}, and instance relationships~\cite{peng2019correlation,tung2019similarity} as the transferred knowledge, which can achieve satisfactory results on various datasets and different DNNs. However, they are usually ineffective in practice when the original data is unusable.

To solve this problem, data-free knowledge distillation~\cite{chawla2021data,zhang2021data} employs synthetic data or webly collected data to train student network with the help of the pre-trained teacher network, which can bypass privacy issues and save data management costs in practical applications. Inspired by the Generative Adversarial Networks~\cite{goodfellow2020generative}, a series of works~\cite{chen2019data,fang2019data,micaelli2019zero} treat the teacher network as the discriminator to supervise a generator to produce pseudo data from random noise. Besides, DeepInversion~\cite{yin2020dreaming} extracts the means and variances stored in the batch normalization layers of teacher network to reconstruct training images. Recently, Contrastive Model Inversion (CMI)~\cite{fang2021contrastive} argues that the instances generated by DeepInversion are highly similar, which is ineffective for student network training. Consequently, CMI augments the diversity of generated data via contrastive learning~\cite{chen2020simple}. Lately, Zhao et al.~\cite{zhao2022dual} use the means and variances of teacher network to guide the generator and further produce new realistic data, thereby improving the performance of student network.

Instead of generating new data for approximating the original data, it is promising to train a satisfactory student network by utilizing the plentiful realistic instances on the Internet. Xu et al.~\cite{xu2019positive} select useful examples from the webly collected data based on a portion of the original data. Chen et al.~\cite{chen2021learning} propose to select useful instances with a low cross-entropy value to train student network. However, they neglect the distribution discrepancies between the webly collected data and original data, which inevitably corrupts the performance of the student network. In this work, we carefully consider and effectively process the distribution shift, thus obtaining a reliable student network.
\subsection{Learning under Distribution Shift}
\label{sec_related_luds}
In the learning scenarios with distribution shift, the training data and test data may come from different distributions ~\cite{pan2010survey,quinonero2008dataset}. In this case, DNNs are biased to training data and cannot perform well during the test phase. To tackle this problem, a series of works~\cite{berthelotremixmatch,wang2019semi} propose to select instances that are similar to the target distribution, and those selected instances are used for retraining the DNNs. Some other works ~\cite{fang2020rethinking,sugiyama2007covariate} adapt the reweighting technique to find out the useful training instances which have the similar distribution to the original data. Furthermore, domain adaption approaches~\cite{du2021generation,zhang2022divide} are proposed to transfer knowledge from the training data (\textit{i.e.}, the source data) to test data (\textit{i.e.}, the target data) , thereby improving the generalization ability of the model under distribution shift.

In data-free knowledge distillation, the distribution of webly collected data is usually different from the unseen test data, which may drop the performance of student network significantly. Therefore, we propose the new method KD$^{3}$ to explicitly deal with such a distribution shift issue for data-free knowledge distillation.
\begin{figure*}[t]
	\centering
	\includegraphics[scale=0.16]{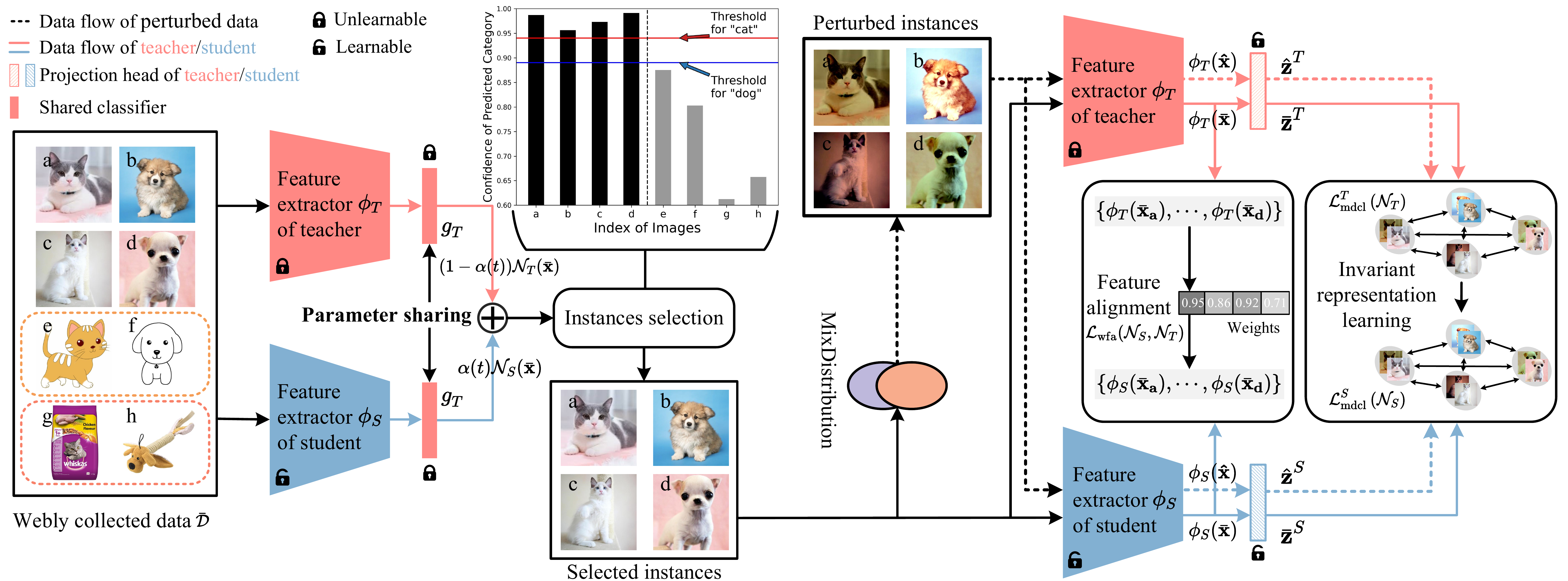}
	\caption{The diagram of our \textbf{K}nowledge \textbf{D}istillation between \textbf{D}ifferent \textbf{D}istributions (KD$^{3}$). The frozen teacher network $\mathcal{N}_{T}$ consists of a feature extractor $\phi_{T}$ and a classifier $g_{T}$. The student network $\mathcal{N}_{S}$ has a learnable feature extractor $\phi_{S}$ and shares $g_{T}$ with $\mathcal{N}_{T}$, where $g_{T}$ is fixed to preserve the information learned by $\mathcal{N}_{T}$. Firstly, the webly collected data $\bar{\mathcal{D}}$ is dynamically selected by $\mathcal{N}_{T}$ and $\mathcal{N}_{S}$. For example, here we assume the original data contain two classes including: ``cat" and ``dog". Then we select the images ``a-d" while discarding the images ``e-h" (with different distributions). Subsequently, the weighted feature alignment conducted on the selected data further promotes $\mathcal{N}_{S}$ to make accurate predictions. Moreover, the MixDistribution contrastive learning is applied to both the perturbed instances (obtained by the MixDistribution) and unperturbed instances, promoting $\mathcal{N}_{S}$ to learn robust representations that are invariant to distribution shift.}
	\label{fig_ts}
\vspace{-1em}
\end{figure*}
\section{Our Approach}
\label{sec_ourap}
In this section, we first introduce some necessary preliminary knowledge, and then we state our KD$^{3}$ on how to learn student networks without using the original data.
\subsection{Preliminary}
Conventional knowledge distillation methods~\cite{hinton2015distilling,romero2014fitnets} seek to learn a small student network $\mathcal{N}_{S}$ by promoting it to mimic the output of a large pre-trained teacher network $\mathcal{N}_{T}$. Formally, we denote the original training data as $\mathcal{D}\!=\!\left\{\left(\mathbf{x}_i, y_i\right)\right\}_{i=1}^{|\mathcal{D}|} \subset \mathcal{X} \times \mathcal{Y}$, where ``$|\cdot|$" is the data cardinality; $\mathcal{X} \subset \mathbb{R}^{I}$ ($I$ is the data dimensionality) and $\mathcal{Y} = \{1,\cdots,K\}$ ($K$ is the total number of classes) are the sample space and label space, respectively. For a training dataset $\mathcal{D}$, the knowledge distillation is accomplished by minimizing the following loss function:
\begin{equation}
\mathcal{L}_{\text{kd}}(\mathcal{N}_{S})=\frac{1}{|\mathcal{D}|} \sum_{i=1}^{|\mathcal{D}|} [\mathcal{H}_{\text{ce}}\left(\mathcal{N}_S\left(\mathbf{x}_i\right), y_i\right)+\lambda\mathcal{H}_{\text{kt}}\left(\mathbf{f}_i^S, \mathbf{f}_i^T\right)],
\end{equation}
where $\mathcal{H}_{\text{ce}}$ is the cross-entropy loss function, encouraging the prediction of student network to be as consistent as the ground-truth; $\mathcal{H}_{\text{kt}}$ is the knowledge transfer function to promote student network to learn the knowledge $\mathbf{f}_i^T$ of teacher network (\textit{e.g.}, predictions or feature maps); $\mathbf{f}_i^S$ is the corresponding knowledge of student network; $\lambda>0$ denotes the trade-off parameter, which is used to balance $\mathcal{H}_{\text{ce}}$ and $\mathcal{H}_{\text{kt}}$.

The necessary original data $\mathcal{D}$ of conventional knowledge distillation methods is usually untouchable due to practical limitations discussed in Section~\ref{sec_intro}. Consequently, a sequence of data-free methods~\cite{chen2021learning,chen2019data,choi2020data} propose to generate the pseudo data from teacher network $\mathcal{N}_{T}$, but the visual quality and diversity of the synthetic images limit their performance. Instead of generating pseudo data, there are massive realistic data $\bar{\mathcal{D}}\!=\!\left\{\left(\bar{\mathbf{x}}_i, \bar{y}_i\right)\right\}_{i=1}^{|\bar{\mathcal{D}}|} \subset \bar{\mathcal{X}} \times \bar{\mathcal{Y}}$ on the Internet which can be gathered to train the student network $\mathcal{N}_{S}$~\cite{chen2021distilling}. Here, the notations with superscript ``--" denote that they are related to the webly collected data. However, there is distribution shift between the webly collected data $\bar{\mathcal{D}}$ and original data $\mathcal{D}$, namely: 1) $p(y|\mathbf{x})\neq p(\bar{y}|\bar{\mathbf{x}})$, \textit{i.e.}, $\bar{\mathcal{D}}$ may contain many uninterested instances due to $\mathcal{Y} \subset \bar{\mathcal{Y}}$ and $|\mathcal{D}| \ll |\bar{\mathcal{D}}|$; 2) $p(\mathbf{x}) \neq p(\bar{\mathbf{x}})$, \textit{i.e.}, the image quality or style of $\mathcal{D}$ and $\bar{\mathcal{D}}$ are different from each other because the instances in $\bar{\mathcal{D}}$ are roughly collected from the Internet. In this case, the student network trained on $\bar{\mathcal{D}}$ inevitably performs poorly on the unseen test data due to the distribution shift. 

To address the aforementioned issue, we propose a novel data-free knowledge distillation method called KD$^{3}$ to train a reliable $\mathcal{N}_{S}$ on the webly collected data $\bar{\mathcal{D}}$. As illustrated in Fig.~\ref{fig_ts}, our KD$^{3}$ contains three key components (as detailed in Sections~\ref{sec_ourap_1}, \ref{sec_ourap_2}, and \ref{sec_ourap_3}, respectively), including 1) Teacher-student dynamic instance selection, which chooses webly collected instances having the similar distribution of original instances; 2) Classifier sharing \& feature alignment, where the student network and teacher network share their classifier parameters and align their output features; 3) MixDistribution contrastive learning, which promotes the student network to produce consistent representations for both perturbed and unperturbed instances.
\subsection{Teacher-Student Dynamic Instance Selection}
\label{sec_ourap_1}
As mentioned above, there is distribution shift between the webly collected data $\bar{\mathcal{D}}\!=\!\left\{\left(\bar{\mathbf{x}}_i, \bar{y}_i\right)\right\}_{i=1}^{|\bar{\mathcal{D}}|}$ and original data $\mathcal{D}$. Since the teacher network is well-trained on $\mathcal{D}$, it is able to show high confidence levels for those instances in $\bar{\mathcal{D}}$ which have the similar distribution with $\mathcal{D}$. Consequently, we propose to select useful instances from $\bar{\mathcal{D}}$ based on the output probabilities of teacher network $\mathcal{N}_{T}$ and student network $\mathcal{N}_{S}$ to alleviate the distribution shift. 

Specifically, we first input all instances in $\bar{\mathcal{D}}$ into both $\mathcal{N}_{T}$ and $\mathcal{N}_{S}$ to get the corresponding output probabilities:
\begin{equation}
\mathcal{N}_T(\bar{\mathbf{x}}_{i})=g_T(\phi_T(\bar{\mathbf{x}}_{i})), \quad \mathcal{N}_S(\bar{\mathbf{x}}_{i})=g_T(\phi_S(\bar{\mathbf{x}}_{i})),
\end{equation} 
where $\phi_T$ and $\phi_S$ denote the feature extractor of $\mathcal{N}_{T}$ and $\mathcal{N}_{S}$, respectively, and $g_T$ represents the shared classifier learned by $\mathcal{N}_{T}$. 
Then, we combine the predictions $\{\mathcal{N}_{T}(\bar{\mathbf{x}}_{i})\}_{i=1}^{|\bar{\mathcal{D}}|}$ and $\{\mathcal{N}_{S}(\bar{\mathbf{x}}_{i})\}_{i=1}^{|\bar{\mathcal{D}}|}$ by the following criterion:
\begin{equation}
\text{Combine}\left(\bar{\mathbf{x}}_i\right)=\left(1-\alpha(t)\right) \mathcal{N}_T\left(\bar{\mathbf{x}}_i\right)+\alpha(t) \mathcal{N}_S\left(\bar{\mathbf{x}}_i\right).
\label{eq_nts}
\end{equation}
The combination is dynamically adjusted by the following time-dependent function:
\begin{equation}
\alpha(t)=\left\{\begin{array}{cl}
\exp \left(-5\left({\frac{t}{I/2}}-1\right)^2\right), & t \leq {I/2}, \\
1, & t> {I/2},
\end{array}\right.
\label{eq_at}
\end{equation}
where $\alpha(t)$ grows from 0 to 1 according to current epoch $t$, and $I$ represents the total number of iterations in training student network. In the early-staged training, the initialized $\mathcal{N}_{S}$ is unable to offer accurate predictions for the instances in $\bar{\mathcal{D}}$, while the pre-trained $\mathcal{N}_{T}$ can precisely recognize the instances in $\bar{\mathcal{D}}$ which are similar to $\mathcal{D}$. Therefore, function $\alpha(t)$ attributes a big weight to $\mathcal{N}_{T}\left(\bar{\mathbf{x}}_i\right)$ at the early stage. With the improvement of $\mathcal{N}_{S}$, $\alpha(t)$ will gradually highlight the importance of $\mathcal{N}_{S}\left(\bar{\mathbf{x}}_i\right)$. When $\alpha(t)=1$ (namely $t\geq{I/2}$), the selection of training instances will be completely determined by the student network $\mathcal{N}_{S}$.

Subsequently, we can obtain the predicted label $y_{i}^{\text{pred}}$ of image $\bar{\mathbf{x}}_{i}$ and the confidence $p_{i}$ of $\bar{\mathbf{x}}_{i}$ belong to $y_{i}^{\text{pred}}$ as:
\begin{equation}
\begin{aligned}
y_{i}^{\text{pred}}&=\arg \max _j\left(\text{Combine}\left(\bar{\mathbf{x}}_i\right)\right)_j, \\ p_{i}&=(\text{Combine}\left(\bar{\mathbf{x}}_i\right))_{y_{i}^{\text{pred}}},
\end{aligned}
\label{eq_yandc}
\end{equation}
respectively. Based on the above $\{y_{i}^{\text{pred}}\}_{i=1}^{\bar{|\mathcal{D}|}}$, we can count the number of labels for each class as $\{n_{i}\}_{i=1}^{K}$, and then we obtain the thresholds $\{\mathcal{T}_{i}\}_{i=1}^{K}$ for filtering out the low-confidence instances of each category as:
\begin{equation}
\mathcal{T}_{i}=\text{Normalization}(n_i) \cdot V_{\text{th}}, 
\label{eq_thr}
\end{equation}
where $V_{\text{th}}$ is a fixed threshold, and $\{n_{i}\}_{i=1}^{K}$ is normalized to $[0,1]$ via the following rule:
\begin{equation}
	\text{Normalization}(n_i)=\frac{n_i}{\max _{1 \leq j \leq K}\left(n_j\right)}.
\end{equation}
Finally, we obtain the useful data $\bar{\mathcal{D}}_{s}\!\!\!=\!\!\!\text{Select}(\bar{\mathbf{x}}_{i})$ (${1\leq i \leq\bar{|\mathcal{D}|}}$) which has the similar distribution with $\mathcal{D}$. Here, the selection operator is defined as:
\begin{equation}
\text{Select}\left(\bar{\mathbf{x}}_i\right)= \begin{cases}\bar{\mathbf{x}}_i \in \bar{\mathcal{D}}_{s}, & p_i>\mathcal{T}_{y_{i}^{\text{pred}}}, \\ \bar{\mathbf{x}}_i \notin \bar{\mathcal{D}}_{s}, & p_i \leq \mathcal{T}_{y_{i}^{\text{pred}}}.\end{cases}
\label{eq_sel}
\end{equation}
During the training phase, $\bar{\mathcal{D}}_{s}$ is continuously updated based on $\mathcal{N}_{T}$, $\mathcal{N}_{S}$, and $\alpha(t)$. If $\mathcal{N}_{S}$ performs worse on a certain category, it will produce low confidence values for the instances of this category and lead to a low threshold. In this case, many instances belonging to this category can be selected to supplement the training of $\mathcal{N}_{S}$ for this category.
\subsection{Classifier Sharing \& Feature Alignment}
\label{sec_ourap_2}
In Section~\ref{sec_ourap_1}, we successfully select the useful data $\bar{\mathcal{D}}_{s}\!=\!\{(\bar{\mathbf{x}}_{i}, \bar{y}_{i})\}_{i=1}^{|\bar{\mathcal{D}}_{s}|}$ from the webly collected data $\bar{\mathcal{D}}$. However, teacher network $\mathcal{N}_{T}$ and student network $\mathcal{N}_{S}$ are unable to make completely correct predictions for all instances in $\bar{\mathcal{D}}$ because both two networks are imperfect, especially in the early-staged iterations. Therefore, the distribution $p(\bar{y}|\bar{\mathbf{x}})$ of $\bar{\mathcal{D}}_{s}$ is still different from $p(y|\mathbf{x})$ of $\mathcal{D}$ in some cases, which may incur inaccurate supervisions to hurt the performance of student network. Recent works~\cite{du2021generation,liang2020we} revealed that the classifiers of DNNs can learn task-specific information. Inspired by this, we share the classifier $g_{T}$ (learned by $\mathcal{N}_{T}$) with $\mathcal{N}_{S}$, so that the critical information of unseen original data (contained in $\mathcal{D}$) can be transferred from $\mathcal{N}_{T}$ to $\mathcal{N}_{S}$. Furthermore, we freeze $g_{T}$ to prevent the information learned from the original data being disturbed by parameter update, which means that $\mathcal{N}_{S}$ only updates its parameters in $\phi_S$ during training. Subsequently, we utilize $\bar{\mathcal{D}}_{s}$ to drive feature alignment between $\mathcal{N}_{S}$ and  $\mathcal{N}_{T}$ in the preceding layer of the shared $g_{T}$, so that the student network can memorize critical knowledge of teacher network as much as possible. 

In detail, the overall goal of our feature alignment is to encourage $\mathcal{N}_{S}$ to produce outputs as consistent as that of $\mathcal{N}_{T}$. Accordingly, we estimate the feature alignment weight $w_i$ for $\bar{\mathbf{x}}_{i} \in \bar{\mathcal{D}}_{s}$ by calculating the consistency between $\mathcal{N}_{T}(\bar{\mathbf{x}}_{i})$ and $\mathcal{N}_{S}(\bar{\mathbf{x}}_{i})$, which is:
\begin{equation}
w_i=1-\operatorname{Sigmoid}\left(\left\|\mathcal{N}_S\left(\bar{\mathbf{x}}_{i}\right)-\mathcal{N}_T\left(\bar{\mathbf{x}}_{i}\right)\right\|_1\right).
\label{eq_weight}
\end{equation}
For an image $\bar{\mathbf{x}}_{i}$, if $\mathcal{N}_{S}$ produces consistent outputs with that of $\mathcal{N}_{T}$, we regard it as an easily-aligned instance and give it a large weight to highlight its positive influence in feature alignment and vice versa. Based on $\{w_i\}_{i=1}^{|\bar{\mathcal{D}}_{s}|}$, the \textbf{w}eighted \textbf{f}eature \textbf{a}lignment loss $\mathcal{L}_{\text{wfa}}(\mathcal{N}_{S}, \mathcal{N}_{T})$ is formulated as:
\begin{equation}
\mathcal{L}_{\text{wfa}}(\mathcal{N}_{S}, \mathcal{N}_{T})\!=\!\!\frac{1}{|\bar{\mathcal{D}}_{s}|}\! \sum_{i=1}^{|\bar{\mathcal{D}}_{s}|} \!w_i \mathcal{H}_{\text{mse}}\!\left(\phi_{S}(\bar{\mathbf{x}}_{i}), \phi_{T}(\bar{\mathbf{x}}_{i})\right),
\label{eq_wfa}
\end{equation}
where $\mathcal{H}_{\text{mse}}$ is the mean square error and it measures the similarity between $\phi_{T}(\bar{\mathbf{x}}_{i})$ and $\phi_{S}(\bar{\mathbf{x}}_{i})$.

Classifier sharing and feature alignment successfully address the shortage of supervision in the student network, thereby eliminating the negative impact of inaccurate labels caused by the webly collected data. When evaluated on the test instance $\mathbf{x}_{\text{test}}$, the student network well aligned with the teacher network can produce feature $\phi_{S}(\mathbf{x}_{\text{test}})$ which is consistent to $\phi_{T}(\mathbf{x}_{\text{test}})$. After that, the parameter-shared classifier $g_{T}$ can produce an accurate prediction $g_{T}(\phi_{S}(\mathbf{x}_{\text{test}}))$ like the teacher prediction $g_{T}(\phi_{T}(\mathbf{x}_{\text{test}}))$.
\subsection{MixDistribution Contrastive Learning}
\label{sec_ourap_3}
In our problem setting, the original training data $\mathcal{D}$ of teacher network $\mathcal{N}_{T}$ is inaccessible and student network $\mathcal{N}_{S}$ trained on the selected data $\bar{\mathcal{D}}_{s}$ needs to correctly recognize the unseen test data. In practice, the original data $\mathcal{D}$ is usually selected and processed manually, so the distribution $p(\bar{\mathbf{x}})$  of webly collected instances cannot accurately match the distribution $p(\mathbf{x})$ of the original data. Recent studies~\cite{yin2020dreaming,zhou2021domain} find that the data distribution is closely related to image style and quality, which can be reflected in statistical variables, \textit{e.g.}, the standard deviation and mean. Therefore, we propose MixDistribution to construct the perturbed data with new distribution, which disturbs statistics of images in $\bar{\mathcal{D}}_{s}$. Finally, we promote student network to learn representation that is invariant to distribution shift by improving the consistency between perturbed and unperturbed instances. 
\begin{table*}[ht]
	\begin{center}
		\resizebox{\linewidth}{!}{
			\begin{tabular}{c|ccccc|ccccccccccccc}
				\hline
				\rule{0pt}{10pt}
				\multirow{1}{*}{Dataset}       & \multirow{1}{*}{Arch} & \multirow{1}{*}{\#params$^{T}$} & \multirow{1}{*}{\#params$^{S}$} & \multirow{1}{*}{FLOPs$^{T}$} & \multirow{1}{*}{FLOPs$^{S}$} & \multicolumn{1}{c}{$\text{ACC}^{T}$} & \multicolumn{1}{c|}{$\text{ACC}^{S}$}    & \multicolumn{1}{c}{DAFL}   & \multicolumn{1}{c}{DFAD}  & \multicolumn{1}{c}{DDAD}  & \multicolumn{1}{c}{DI}    & \multicolumn{1}{c}{ZSKT} & \multicolumn{1}{c}{PRE}  & \multicolumn{1}{c}{DFQ}   & \multicolumn{1}{c}{CMI}    & \multicolumn{1}{c|}{DFND}   & \multicolumn{1}{c}{KD$^{3}$}   & \multicolumn{1}{c}{ACC${\uparrow}$}   \\ \hline
				\multirow{1}{*}{MNIST}       &   $\nabla$  &0.062M &0.019M &0.42M &0.14M & \multicolumn{1}{c}{98.91}   & \multicolumn{1}{c|}{98.65}       & \multicolumn{1}{c}{98.20} & \multicolumn{1}{c}{98.31} & \multicolumn{1}{c}{98.09} & \multicolumn{1}{c}{--} & \multicolumn{1}{c}{97.44}   & \multicolumn{1}{c}{98.33}   & \multicolumn{1}{c}{97.49} & \multicolumn{1}{c}{--} & \multicolumn{1}{c|}{$\underline{\text{98.37}}$}  & \multicolumn{1}{c}{\textbf{98.76}} & +\textcolor{green}{0.39} \\ \hline
				\multirow{2}{*}{CIFAR10}       &   $\diamondsuit$  &21.28M &11.17M &1.16G &0.56G  & \multicolumn{1}{c}{95.70}   & \multicolumn{1}{c|}{95.20}       & \multicolumn{1}{c}{92.22} & \multicolumn{1}{c}{93.30} & \multicolumn{1}{c}{93.08} & \multicolumn{1}{c}{93.26} & \multicolumn{1}{c}{93.32}   & \multicolumn{1}{c}{93.25}   & \multicolumn{1}{c}{94.61} & \multicolumn{1}{c}{$\underline{\text{94.84}}$} & \multicolumn{1}{c|}{94.02}  & \multicolumn{1}{c}{\textbf{95.21}} & +\textcolor{green}{0.37} \\ 
				&    $\heartsuit$    &14.73M &9.42M &0.40G &0.28G    & \multicolumn{1}{c}{94.07}  & \multicolumn{1}{c|}{92.69}           & \multicolumn{1}{c}{86.92} & \multicolumn{1}{c}{90.38} & \multicolumn{1}{c}{90.85} & \multicolumn{1}{c}{85.27} & \multicolumn{1}{c}{91.22} & \multicolumn{1}{c}{91.82}& \multicolumn{1}{c}{91.36} & \multicolumn{1}{c}{88.49}  & \multicolumn{1}{c|}{$\underline{\text{92.61}}$} & \multicolumn{1}{c}{\textbf{94.13}}  & +\textcolor{green}{1.52} \\   \hline
				\multirow{2}{*}{CIFAR100}      &  $\diamondsuit$   &21.28M &11.17M &1.16G &0.56G    & \multicolumn{1}{c}{78.05}    & \multicolumn{1}{c|}{77.10}              & \multicolumn{1}{c}{74.47} & \multicolumn{1}{c}{67.70} & \multicolumn{1}{c}{73.64} & \multicolumn{1}{c}{61.32} & \multicolumn{1}{c}{67.74} & \multicolumn{1}{c}{74.19}       & \multicolumn{1}{c}{77.01} & \multicolumn{1}{c}{$\underline{\text{77.04}}$} & \multicolumn{1}{c|}{76.35}  & \multicolumn{1}{c}{\textbf{78.44}} & +\textcolor{green}{1.40} \\  
				&   $\heartsuit$  &14.73M &9.42M &0.40G &0.28G    & \multicolumn{1}{c}{74.53}    & \multicolumn{1}{c|}{72.28}            & \multicolumn{1}{c}{65.36} & \multicolumn{1}{c}{64.90} & \multicolumn{1}{c}{68.33} & \multicolumn{1}{c}{60.00} & \multicolumn{1}{c}{58.33} & \multicolumn{1}{c}{70.34}        & \multicolumn{1}{c}{62.53} & \multicolumn{1}{c}{59.70}  & \multicolumn{1}{c|}{$\underline{\text{70.88}}$} & \multicolumn{1}{c}{\textbf{74.21}} & +\textcolor{green}{3.33} \\  \hline
				\multirow{2}{*}{CINIC}         &  $\diamondsuit$  &21.28M &11.17M &1.16G &0.56G    & \multicolumn{1}{c}{86.62}   & \multicolumn{1}{c|}{85.09}              & \multicolumn{1}{c}{60.54} & \multicolumn{1}{c}{71.38} & \multicolumn{1}{c}{80.10} & \multicolumn{1}{c}{78.57} & \multicolumn{1}{c}{64.73} & \multicolumn{1}{c}{77.56}      & \multicolumn{1}{c}{71.76} & \multicolumn{1}{c}{78.47}  & \multicolumn{1}{c|}{$\underline{\text{82.96}}$} & \multicolumn{1}{c}{\textbf{86.55}}  & +\textcolor{green}{3.59} \\ 
				&    $\heartsuit$  &14.73M &9.42M &0.40G &0.28G    & \multicolumn{1}{c}{84.22}   & \multicolumn{1}{c|}{83.28}           & \multicolumn{1}{c}{59.08} & \multicolumn{1}{c}{60.67} & \multicolumn{1}{c}{77.90} & \multicolumn{1}{c}{68.90} & \multicolumn{1}{c}{58.84} & \multicolumn{1}{c}{65.38}      & \multicolumn{1}{c}{74.33} & \multicolumn{1}{c}{74.99}  & \multicolumn{1}{c|}{$\underline{\text{81.82}}$} & \multicolumn{1}{c}{\textbf{83.54}}   & +\textcolor{green}{1.72}  \\   \hline
				\multirow{2}{*}{TinyImageNet} &  $\diamondsuit$  &21.28M &11.17M &4.65G &2.23G  & \multicolumn{1}{c}{66.44} & \multicolumn{1}{c|}{64.87}
				& \multicolumn{1}{c}{52.20}      & \multicolumn{1}{c}{20.63}      & \multicolumn{1}{c}{59.84}      & \multicolumn{1}{c}{6.98}      & \multicolumn{1}{c}{31.51}   & \multicolumn{1}{c}{50.15}       & \multicolumn{1}{c}{63.73}      & \multicolumn{1}{c}{$\underline{\text{64.01}}$}       & \multicolumn{1}{c|}{60.92}       & \multicolumn{1}{c}{\textbf{66.24}}      & +\textcolor{green}{2.23}    \\  
				&  $\heartsuit$   &14.73M &9.42M &1.26G &0.92G     & \multicolumn{1}{c}{62.34}   & \multicolumn{1}{c|}{61.55}            & \multicolumn{1}{c}{53.89}      & \multicolumn{1}{c}{38.95}      & \multicolumn{1}{c}{42.25}      & \multicolumn{1}{c}{1.22}      & \multicolumn{1}{c}{30.63}     & \multicolumn{1}{c}{45.92}    & \multicolumn{1}{c}{23.43}      & \multicolumn{1}{c}{17.73}       & \multicolumn{1}{c|}{$\underline{\text{56.87}}$}       & \multicolumn{1}{c}{\textbf{61.98}}    & +\textcolor{green}{5.11}    \\ \hline
		\end{tabular}}
	\end{center}
	\vspace{-0.5em}
	\caption{Classification accuracy (in \%) of the student network trained by various methods on five image classification datasets. The notations $\nabla$, $\diamondsuit$, and $\heartsuit$ represent the teacher-student pairs LeNet5-LeNet5\_half, ResNet34-ResNet18, and VGGNet16-VGGNet13, respectively. $\text{ACC}$, \#params, and floating point operations (FLOPs) denote the yielded accuracy, parameters (in millions, M), and calculations (in Gigas, G) of the corresponding DNN, respectively. These notations with superscripts ``$T$" and ``$S$" represent that they are related to the teacher network and student network, respectively. The best results achieved by baseline methods are underlined, and the column ``ACC${\uparrow}$" with green fonts shows the accuracy improvement of KD$^{3}$ in contrast to the best results among compared baseline methods.}
	\label{table_class}
	\vspace{-1em}
\end{table*}

More specifically, we first randomize $\{\bar{\mathbf{x}}_{i}\}_{i=1}^{|\bar{\mathcal{D}}_{s}|}$ as $\text {Randomize }(\{\bar{\mathbf{x}}_{i}\}_{i=1}^{|\bar{\mathcal{D}}_{s}|})$ and compute the perturbed statistics by the following rules:
\begin{equation}
\left\{\begin{array}{l}
\gamma_{\text {mix}}=\lambda \sigma(\{\bar{\mathbf{x}}_{i}\}_{i=1}^{|\bar{\mathcal{D}}_{s}|})+(1-\lambda) \sigma(\text{Rand}(\{\bar{\mathbf{x}}_{i}\}_{i=1}^{|\bar{\mathcal{D}}_{s}|})),\\
\beta_{\text {mix}}=\lambda \mu(\{\bar{\mathbf{x}}_{i}\}_{i=1}^{|\bar{\mathcal{D}}_{s}|})+(1-\lambda) \mu(\text{Rand}(\{\bar{\mathbf{x}}_{i}\}_{i=1}^{|\bar{\mathcal{D}}_{s}|})),
\end{array}\right.
\end{equation}
where $\lambda>0$ is produced by beta distribution $\operatorname{Beta}(\delta, \delta)$ with $\delta \in(0, \infty)$ being a hyper-parameter.  Here, $\sigma(\cdot)$ and $\mu(\cdot)$ denote the standard deviation and mean of the corresponding variables, respectively. Then, we construct the perturbed image $\hat{\mathbf{x}}_{i}$ by:
\begin{equation}
\hat{\mathbf{x}}_{i}=\gamma_{\text{mix}} \frac{\bar{\mathbf{x}}_{i}-\mu(\bar{\mathbf{x}}_{i})}{\sigma(\bar{\mathbf{x}}_{i})}+\beta_{\text{mix}},
\label{eq_mix}
\end{equation}
where we scale and shift the normalized $\bar{\mathbf{x}}_{i}$ by $\gamma_{\text{mix}}$ and $\beta_{\text{mix}}$, respectively. After the above instance perturbation, the raw images $\{\bar{\mathbf{x}}_{i}\}_{i=1}^{|\bar{\mathcal{D}}_{s}|}$ and the perturbed images $\{\hat{\mathbf{x}}_{i}\}_{i=1}^{|\bar{\mathcal{D}}_{s}|}$ are fed into $\mathcal{N}_{T}$ and $\mathcal{N}_{S}$ to obtain the features in penultimate layer. Subsequently, we follow~\cite{chen2020simple} to transfer features of all dimensionalities into the embedding space by a projection head. By taking the teacher network $\mathcal{N}_{T}$ and the corresponding feature  $\phi_{T}(\bar{\mathbf{x}}_{i})$ as an example, the embedding result $\bar{\mathbf{z}}_i^T$ is calculated by:
\begin{equation}
\bar{\mathbf{z}}_i^T=\text { Normalization }\left(\mathbf{W}_p^T \phi_{T}({\bar{\mathbf{x}}}_{i})+\mathbf{b}_p^T\right),
\label{eq_map}
\end{equation}
where $\mathbf{W}_p^T$ and $\mathbf{b}_p^T$ denote the weight and bias of projection head, and the notations with superscripts ``$T$" and ``$S$" represent they are related to teacher network and student network, respectively. Similarly, the embedding result $\hat{\mathbf{z}}_i^T$ of perturbed example $\hat{\mathbf{x}}_{i}$ is computed by:
\begin{equation}
\hat{\mathbf{z}}_i^T=\text { Normalization }\left(\mathbf{W}_p^T \phi_{T}({\hat{\mathbf{x}}}_{i})+\mathbf{b}_p^T\right).
\label{eq_map_hat}
\end{equation}
Based on the embeddings of unperturbed instances $\{\bar{\mathbf{z}}_i^T\}_{i=1}^{|\bar{\mathcal{D}}_{s}|}$ and perturbed instances $\{\hat{\mathbf{z}}_i^T\}_{i=1}^{|\bar{\mathcal{D}}_{s}|}$, we can calculate the \textbf{M}ix\textbf{D}istribution \textbf{c}ontrastive \textbf{l}earning (MDCL) loss $\mathcal{L}_{\text {mdcl}}^{T}(\mathcal{N}_T)$  for the teacher network as follows: 
\begin{equation}
\!\mathcal{L}_{\text {mdcl}}^{T}(\mathcal{N}_T)\!\!=\!\!-\sum_{i=1}^{|\bar{\mathcal{D}}_{s}|}\! \log \frac{\exp \left(\operatorname{sim} \left(\bar{\mathbf{z}}_i^T, \hat{\mathbf{z}}_i^T\right) / \tau\right)}{\!\sum_{j=1}^{ |\bar{\mathcal{D}}_{s}|} \mathbbm{1}_{[j \neq i]} \exp \left(\operatorname{sim}\left(\bar{\mathbf{z}}_i^T, \hat{\mathbf{z}}_j^T\right) / \tau\right)\!},
\label{eq_mdcl_t}
\end{equation}
where $\tau>0$ is a temperature parameter; $\operatorname{sim}(\cdot)$ denotes the well-known cosine similarity~\cite{wojke2018deep}; $\mathbbm{1}_{[j \neq i]}$ is the indicator function, and its value is 0 only if $i=j$, and its value is 1, otherwise. Likewise, the MDCL loss of student network is:
\begin{equation}
\!\mathcal{L}_{\text {mdcl}}^{S}(\mathcal{N}_S)\!\!=\!-\sum_{i=1}^{|\bar{\mathcal{D}}_{s}|}\! \log \frac{\exp \left(\operatorname{sim}\! \left(\bar{\mathbf{z}}_i^S, \hat{\mathbf{z}}_i^S\right) / \tau\right)}{\!\sum_{j=1}^{ |\bar{\mathcal{D}}_{s}|} \mathbbm{1}_{[j \neq i]} \exp \left(\operatorname{sim}\left(\bar{\mathbf{z}}_i^S, \hat{\mathbf{z}}_j^S\right) / \tau\right)\!}.
\label{eq_mdcl_s}
\end{equation}
$\mathcal{L}_{\text {mdcl}}^{T}(\mathcal{N}_T)$ and $\mathcal{L}_{\text {mdcl}}^{S}(\mathcal{N}_S)$ depict the relationship among the embeddings of $\mathcal{N}_{T}$ and $\mathcal{N}_{S}$, respectively. Recent studies~\cite{peng2019correlation,tung2019similarity,zhu2021complementary} have demonstrated that transferring the relationship between representations is more effective than transferring representations directly. Therefore, by following~\cite{zhu2021complementary}, we integrate the following learning objectives of $\mathcal{N}_{T}$ and $\mathcal{N}_{S}$ based on the similarity relationship:
\begin{equation}
\mathcal{L}_{\text {mdcl}}(\mathcal{N}_{S},\mathcal{N}_{T})=\mathcal{L}_{\text {mdcl}}^{S}(\mathcal{N}_{S}) + \mathcal{L}_{\text {mdcl}}^{T}(\mathcal{N}_{T}).
\label{eq_mdcl}
\end{equation}

By minimizing $\mathcal{L}_{\text {mdcl}}(\mathcal{N}_{S},\mathcal{N}_{T})$, the student network is encouraged to produce close representations for perturbed and unperturbed versions of the same instance, despite distribution shift between the two versions. Therefore, the student network can accurately classify the test instances of which the distribution is different from $\bar{\mathcal{D}}_{s}$. Note that the parameters in the projection heads of teacher and student will keep updating during the training phase. 

\textbf{Overall Learning Objective}. In the end, the complete objective function of our KD$^{3}$ is:
\begin{equation}
\mathcal{L}_{\text {objective }}(\!\mathcal{N}_{S},\mathcal{N}_{T}\!)\!=\!\mathcal{L}_{\text{wfa}}(\!\mathcal{N}_{S},\mathcal{N}_{T}\!)\!+\!\alpha \mathcal{L}_{\text {mdcl}}(\!\mathcal{N}_{S},\mathcal{N}_{T}\!),
\label{eq_objective}
\end{equation}
where $\alpha$ is a non-negative trade-off parameter to balance the weighted feature alignment loss $\mathcal{L}_{\text{wfa}}(\mathcal{N}_{S},\mathcal{N}_{T})$ and MixDistribution contrastive learning loss $\mathcal{L}_{\text {mdcl}}(\mathcal{N}_{S},\mathcal{N}_{T})$. The detailed training algorithm of KD$^{3}$ is summarized in \textbf{supplementary material}.
\section{Experiments}
\label{sec_exp}
In this section, we demonstrate the effectiveness of our proposed KD$^{3}$ on multiple image classification datasets. 
\textbf{Compared Methods:} We compare our proposed KD$^{3}$ with representative data-free methods, including Data-Free Learning (DAFL)~\cite{chen2019data}, Data-Free Adversarial Learning (DFAD)~\cite{fang2019data}, Dual Discriminator Adversarial Distillation (DDAD)~\cite{zhao2022dual}, DeepInversion (DI)~\cite{yin2020dreaming}, Zero-Shot Knowledge Transfer (ZSKT)~\cite{micaelli2019zero}, Pseudo Replay Enhanced
Data-Free Knowledge Distillation (PRE)~\cite{binici2022robust}, Data-Free Quantization (DFQ)~\cite{choi2020data}, Contrastive Model Inversion (CMI)~\cite{fang2021contrastive}, and Data-Free Noisy Distillation (DFND)~\cite{chen2021learning} (the only existing method using the webly collected data). We implement the above methods by using the codes on their official GitHub pages. 
\\
\textbf{Original Datasets:} We verify our proposed KD$^{3}$ on the test set of MNIST~\cite{lecun1998gradient}, CIFAR10~\cite{krizhevsky2009learning}, CINIC~\cite{darlow2018cinic}, CIFAR100~\cite{krizhevsky2009learning}, and TinyImageNet~\cite{le2015tiny}. \\
\textbf{Webly Collected Datasets:} When using MNIST as the original data, we adopt the training images from both MNIST-M~\cite{ganin2016domain} and SVHN~\cite{netzer2011reading} datasets as the webly collected data, and we grayscale the images in MNIST-M and SVHN because the images in MNIST only have one channel. When using other datasets as the original data, we employ the training images from the large-scale ImageNet~\cite{deng2009imagenet}. We also downsample the images in ImageNet to 32$\times$32 or 64$\times$64 to ensure the size consistency between the original data and webly collected data. Details of the adopted datasets are provided in \textbf{supplementary material}.\\
\textbf{Implementation Details:} When training on MNIST, we use Adam with the initial learning rate of 10$^{-3}$ as the optimizer, and all student networks are trained with 40 epochs. When training on other datasets, we utilize Stochastic Gradient Descent (SGD) with weight decay of 5$\times$10$^{-4}$ and momentum of 0.9 as the optimizer. By following~\cite{chen2022knowledge}, all student networks are trained with 240 epochs, and the initial learning rate is set to 0.05, which is divided by ten at 150, 180, and 210 epochs. Besides, the temperature $\tau$ in Eq.~\eqref{eq_mdcl_t} and Eq.~\eqref{eq_mdcl_s}, threshold $V_{\text{th}}$ in Eq.~\eqref{eq_sel}, and trade-off parameter $\alpha$ in Eq.~\eqref{eq_objective} are 0.30, 0.95, and 0.01, respectively. The parametric sensitivity will be investigated in Section~\ref{exp_ps}.
\begin{table}
	\begin{center}
	\resizebox{\linewidth}{!}{
	\begin{tabular}{c|c|cc}
	\hline
	Operation                                                                                           & Type             & CIFAR10 & CIFAR100 \\ \hline
	\multirow{2}{*}{\begin{tabular}[c]{@{}c@{}}No classifier \\ sharing\end{tabular}}                      & One-hot          & 93.42 ($-$\textcolor{red}{1.79}) & 74.54 ($-$\textcolor{red}{3.90})  \\  
	& Soft             & 93.98 ($-$\textcolor{red}{1.23})  & 76.92 ($-$\textcolor{red}{1.52})  \\ \hline
	\multirow{3}{*}{\begin{tabular}[c]{@{}c@{}}Instance \\ selection\end{tabular}}                       & Random           & 90.22 ($-$\textcolor{red}{4.99})  & 73.60 ($-$\textcolor{red}{4.84})  \\  
	& Only $\mathcal{N}_{S}$          & 91.99 ($-$\textcolor{red}{3.22}) & 75.30 ($-$\textcolor{red}{3.14})  \\ 
	& Only $\mathcal{N}_{T}$           & 94.01 ($-$\textcolor{red}{1.20}) & 76.76 ($-$\textcolor{red}{1.68})  \\ \hline
	\multirow{2}{*}{\begin{tabular}[c]{@{}c@{}}MDCL\end{tabular}} & No $\mathcal{L}_{\text{mdcl}}$             & 94.48 ($-$\textcolor{red}{0.73}) & 77.35 ($-$\textcolor{red}{1.09})  \\  
	& No MD & 94.61 ($-$\textcolor{red}{0.60}) & 77.43 ($-$\textcolor{red}{1.01})  \\ \hline
	\multirow{1}{*}{KD$^{3}$} & $\mathcal{L}_{\text {objective }}$             & \textbf{95.21}  & \textbf{78.44}     \\ \hline
\end{tabular}}
	\end{center}
	\vspace{-0.5em}
	\caption{Classification accuracy (in \%) of ablation experiments. Brackets with red font denote the accuracy drop of the corresponding item compared with the complete KD$^{3}$.}
\label{table_ab}
\vspace{-1.4em}
\end{table}
\subsection{Experiments on Image Classification Datasets}
\label{exp_class}
In this section, we conduct intensive experiments on five image classification tasks mentioned above to demonstrate the effectiveness of our proposed KD$^{3}$. We select four teacher-student pairs for experiments, including LeNet5-LeNet5\_half~\cite{lecun1998gradient}, ResNet34-ResNet18~\cite{he2016deep}, VGGNet16-VGGNet13~\cite{simonyan2014very}, which are widely used in data-free methods~\cite{binici2022robust,chen2019data}. The experimental results are reported in Table~\ref{table_class}.

Firstly, the performance of student networks trained on synthetic data is suboptimal in general, particularly when evaluating on the complex TinyImageNet, because the generated data is usually flawed or distorted. Secondly, we can observe that DFND using the instances on the Internet is still unable to produce a student network competitive to that trained on the original data, which is due to the ignorance of the distribution shift between the webly collected data and original data. In contrast, our KD$^{3}$ can successfully acquire the student networks which achieve significantly better performance than those trained on the original data in most cases. The experimental results demonstrate that our KD$^{3}$ can effectively resolve the distribution shift between the webly collected data and original data, thus training a superior student network without any original data.
\begin{figure}[t]
	\centering
	\includegraphics[scale=0.10]{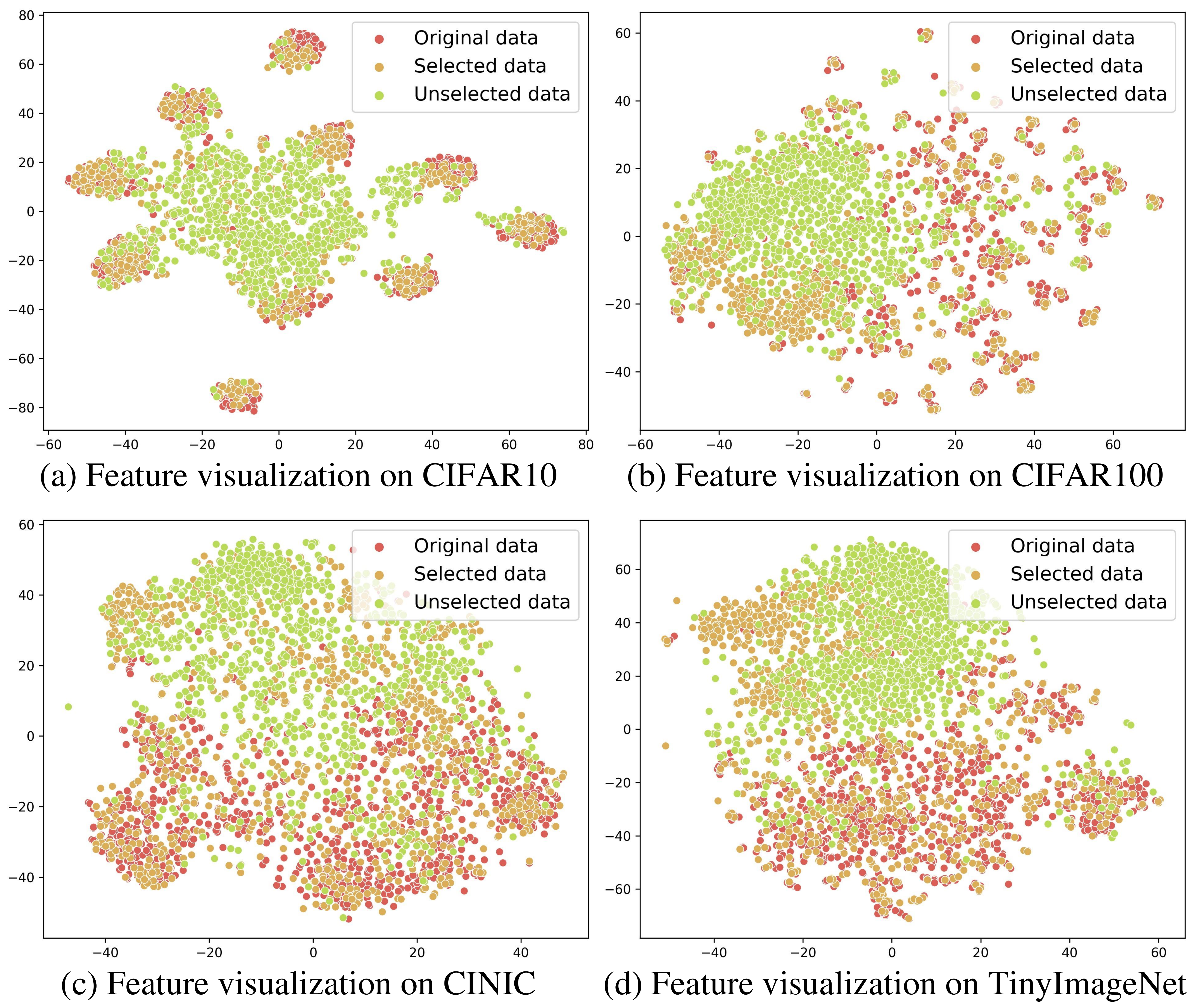}
	\caption{Visualization of ResNet34-produced features by t-SNE~\cite{van2008visualizing}. The original images are from (a) CIFAR10, (b) CIFAR100, (c) CINIC, and (d) TinyImageNet, while the selected and unselected images are from ImageNet, each part contains 1,000 images. The data points selected by our KD$^{3}$ (\textit{i.e.}, orange dots) show very similar distribution with the original data (\textit{i.e.}, red dots).}
	\label{fig_fv}
\vspace{-1.5em}
\end{figure}
\begin{table*}
	\begin{center}
\resizebox{\linewidth}{!}{
\begin{tabular}{cccccccccccccc}
	\hline
	\multicolumn{1}{c}{\multirow{2}{*}{Teacher}} & \multicolumn{1}{c}{\multirow{2}{*}{Student}} & \multicolumn{2}{c}{\#params} & \multicolumn{2}{c}{FLOPs} & \multicolumn{2}{c}{CIFAR10}                      & \multicolumn{2}{c}{CIFAR100}                 & \multicolumn{2}{c}{CINIC}                     & \multicolumn{2}{c}{TinyImageNet}                        \\
	\multicolumn{1}{c}{}                         & \multicolumn{1}{c}{}      & \multicolumn{1}{c}{Teacher} & \multicolumn{1}{c}{Student}  & \multicolumn{1}{c}{Teacher} & \multicolumn{1}{c}{Student}                    & \multicolumn{1}{c}{$\text{ACC}^{S}$} & \multicolumn{1}{c}{KD$^{3}$} & \multicolumn{1}{c}{$\text{ACC}^{S}$} & \multicolumn{1}{c}{KD$^{3}$} & \multicolumn{1}{c}{$\text{ACC}^{S}$} & \multicolumn{1}{c}{KD$^{3}$} & \multicolumn{1}{c}{$\text{ACC}^{S}$} & \multicolumn{1}{c}{KD$^{3}$} \\ \hline
	ResNet32$\times$4    & ResNet8$\times$4   &7.41M &1.21M &1.09G &0.18G & 92.09    & 93.05    & 73.09    & 73.17    & 81.74   & 81.71     & 55.40    & 55.13                 \\
	ResNet32$\times$4    & MobileNetV2        &7.41M &0.81M &1.09G &7.37M & 92.38    & 92.16    & 69.06    & 69.40    & 77.61   & 77.95     & 57.15    & 60.60                 \\
	ResNet32$\times$4    & ShuffleV1          &7.41M &0.86M &1.09G &42.11M & 92.92    & 93.24    & 66.43    & 72.15    & 80.13   & 80.93     & 57.94    & 60.01                 \\
	ResNet32$\times$4    & ShuffleV2          &7.41M &1.26M &1.09G &46.66M & 93.23    & 93.53    & 72.60    & 73.14    & 80.64   & 80.74     & 60.93    & 61.41                 \\
	ResNet110$\times$2   & ResNet110         &6.89M &1.73M &1.02G &0.26G  & 93.37    & 94.59    & 74.31    & 73.59    & 84.29   & 84.75     & 59.80    & 60.21                 \\
	ResNet110$\times$2   & ResNet116         &6.89M &1.83M &1.02G &0.27G  & 93.21    & 94.57    & 74.46    & 73.75    & 84.45   & 84.68     & 59.85    & 59.52                 \\
	ResNet110$\times$2   & ShuffleV1         &6.89M &0.86M &1.02G &42.11M  & 92.92    & 93.24    &66.43     & 72.15    & 80.13   & 81.25     & 57.94    & 58.54                 \\
	ResNet110$\times$2   & ShuffleV2        &6.89M &1.26M &1.02G  &46.66M   & 93.23    & 93.46    & 72.60    & 72.93    & 80.64   & 81.54     & 60.93    &  60.80                 \\ \hline
\end{tabular}}
\end{center}

\caption{Classification accuracy (in \%) of various network backbones. The columns ``$\text{ACC}^{S}$" report the accuracies yielded by the student networks using the original data. Here, the FLOPs are calculated by feeding a 32$\times$32 sized RGB image into the corresponding DNN.}
\label{table_extent}
\end{table*}
\subsection{Ablation Studies \& Feature Visualization}
\label{exp_ab}
\textbf{Ablation Studies.} We select the teacher-student pair ResNet34-ResNet18 to evaluate the three key operations in KD$^{3}$, and the results are shown in Table~\ref{table_ab}. The contributions of these key operations are analyzed as follows:\\
\textbf{1) Classifier Sharing} in Section~\ref{sec_ourap_2}: To estimate the effectiveness of sharing the classifier of teacher with student, we train a student with an initialized classifier. Moreover, to train the initialized classifier, we utilize a cross-entropy or Kullback-Leibler (KL)~\cite{hinton2015distilling} divergence to enforce student network to mimic the one-hot predictions or soft labels of teacher network (shown in ``One-hot'' and ``Soft''). It can be found that the performance of student network with the initialized classifier obviously degrades, which indicates that classifier sharing is vital to enhancing student's performance. It means that our method can effectively transfer the teacher-learned information of original data to student.\\
\textbf{2) Instance Selection} in Section~\ref{sec_ourap_2}: The student network obtains poor performance when the instances are randomly selected (shown in ``Random'') and only selected by student network (shown in ``Only $\mathcal{N}_{S}$''). Furthermore, the student network that trained on the instances chosen by the powerful teacher network achieves relatively good performance (shown in ``Only $\mathcal{N}_{T}$''). In particular, the student network achieves the best accuracy when utilizing the data selected by our proposed data selection method, demonstrating that our proposed data selection method can sample proper instances for student network training.\\
\textbf{3) MixDistribution Contrastive Learning} in Section~\ref{sec_ourap_3}: We directly remove $\mathcal{L}_{\text{mdcl}}(\mathcal{N}_{S}, \mathcal{N}_{T})$ (shown in ``No $\mathcal{L}_{\text{mdcl}}$'') or replace MixDistribution by data augmentations as in~\cite{zhang2017mixup}  (shown in ``No MD'') to train student network. The accuracy of student network has reduced significantly when evaluated on test data of which the distribution is different from the webly collected data. The results demonstrate that MixDistribution contrastive learning is critical to solving the distribution shift problem.

\textbf{Visualization of Features.} To further understand the effectiveness of our data selection method, we visualize the ResNet34-provided features of images from original data, selected data, and unselected data. The original training images are provided by CIFAR10, CIFAR100, CINIC, and TinyImageNet, and webly collected images are from ImageNet. The t-SNE~\cite{van2008visualizing} visualization results are shown in Fig.~\ref{fig_fv}, from which we can observe that the distributions of selected images are close to the original images in feature space. The visualization results demonstrate that our data selection method can effectively select the webly collected instances with the similar distribution to original data. More visualization results are shown in \textbf{supplementary material}.
\subsection{Parametric Sensitivity}
\label{exp_ps}
The tuning parameters in our KD$^{3}$ include the trade-off parameter $\alpha$ in Eq.~\eqref{eq_objective}, temperature parameter $\tau$ in Eq.~\eqref{eq_mdcl_t} and Eq.~\eqref{eq_mdcl_s}, and threshold parameter $V_{\text{th}}$ in Eq.~\eqref{eq_sel}. This section analyzes the sensitivity of our KD$^{3}$ to these parameters on the CIFAR dataset. The ResNet34 and ResNet18 are selected as teacher and student, respectively. We examine the resulting accuracy during training by changing one parameter while holding the others. 

\begin{figure}[t]
	\centering
	\includegraphics[scale=0.13]{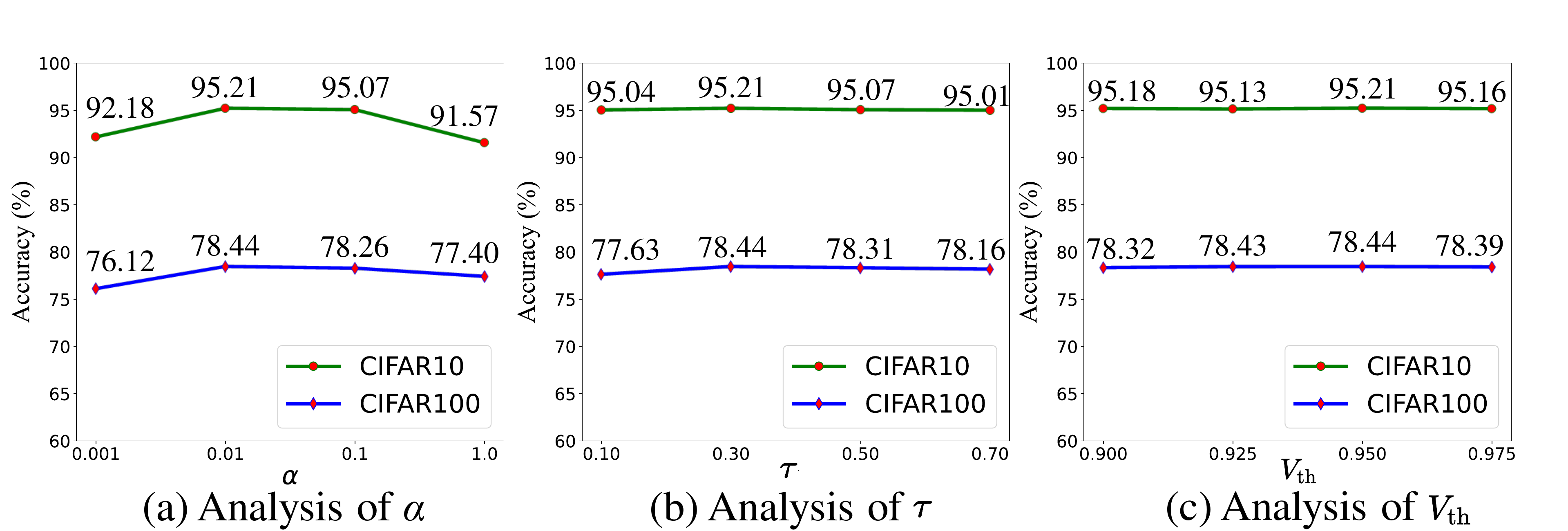}
	\caption{Parametric sensitivity of (a) $\alpha$ in Eq.~\eqref{eq_objective}, (b) $\tau$ in Eq.~\eqref{eq_mdcl_t} and Eq.~\eqref{eq_mdcl_s}, and (c) $V_{\text{th}}$ in Eq.~\eqref{eq_thr}.}
	\label{fig_ps}
\end{figure}

Fig.~\ref{fig_ps} depicts the curves of test accuracy for student network when the parameters vary. The parameters $\alpha$, $\tau$, and $V_{\text{th}}$ vary within $\{0.001, 0.01, 0.1, 1\}$, $\{0.1, 0.3, 0.5, 0.7\}$, and $\{0.900, 0.925, 0.950, 0.975\}$, respectively. Even though these parameters vary over a wide range, we can obverse that the curves of accuracy are generally smooth and relatively stable, which indicates that the performance of student is robust to the variations of parameters. Therefore, the parameters in our KD$^{3}$ are easy to tune.
\vspace{-1em}
\subsection{Experiments with More Network Backbones}
\label{exp_extent}
In this section, we conduct intensive experiments on four benchmark datasets to further verify the performance of KD$^{3}$ equipped with various widely-used teacher-student pairs~\cite{he2016deep,ma2018shufflenet,sandler2018mobilenetv2,zhang2018shufflenet}. The results are reported in Table~\ref{table_extent}. It can be found that the student networks trained by our KD$^{3}$ consistently achieve competitive performance to those trained on the original data, even though some student networks are with different styles of the teacher network. The experimental results demonstrate that our data-free method KD$^{3}$ can be flexibly employed to teacher-student pairs with various structures to train reliable student networks.
\section{Conclusion}
\label{sec_con}
This paper proposed a new data-free approach termed KD$^{3}$ to train student networks using the webly collected data. To our best knowledge, we are the first to address the commonly overlooked yet important distribution shift issue between the webly collected data and original data in knowledge distillation. Our proposed KD$^{3}$ adopts three main techniques to tackle such distribution shift, namely: 1) selection of webly collected instances with the similar distribution to original data; 2) alignment of feature distributions between the teacher network and student network with parameter-shared classifiers; and 3) promotion of feature consistency for input instances and MixDistribution-generated instances. Intensive experiments demonstrated that our KD$^{3}$ can effectively handle the distribution shift to train reliable student networks without using the original training data.\\
\textbf{Acknowledgment}. \hspace{-4mm}C.G. was supported by NSF of China (No: 61973162), NSF of Jiangsu Province (No: BZ2021013), NSF for Distinguished Young Scholar of Jiangsu Province (No: BK20220080), the Fundamental Research Funds for the Central Universities (Nos: 30920032202, 30921013114), CAAI-Huawei MindSpore Open Fund, and “111” Program. M.S. was supported by the Institute for AI and Beyond, UTokyo.
{\small
\bibliographystyle{main}
\bibliography{main}
}
\end{document}